%% file: colm2026_conference.tex
\documentclass{article}
\usepackage[preprint]{colm2026_conference}

\usepackage{microtype}
\usepackage{hyperref}
\usepackage{url}
\usepackage{booktabs}
\usepackage{graphicx}
\usepackage{multirow}
\usepackage{amsmath}
\usepackage{amssymb}
\usepackage{lineno}
\usepackage{pifont}
\usepackage{fontawesome5}
\usepackage{amssymb}
\definecolor{darkblue}{rgb}{0, 0, 0.5}
\hypersetup{colorlinks=true, citecolor=darkblue, linkcolor=darkblue, urlcolor=darkblue}

\newcommand{\best}[1]{\textbf{#1}}
\newcommand{\second}[1]{\underline{#1}}

\title{Select-then-Solve: Paradigm Routing as Inference-Time Optimization for LLM Agents}

\author{
\textbf{Heng Zhou}, \textbf{Zelin Tan}, \textbf{Zhemeng Zhang}, \textbf{Yutao Fan}, \textbf{Yibing Lin}, \textbf{Li Kang}, \\
\textbf{Xiufeng Song}, \textbf{Rui Li}, \textbf{Songtao Huang}, \textbf{Ao Yu}, \textbf{Yuchen Fan}, \textbf{Yanxu Chen}, \\
\textbf{Kaixin Xu}, \textbf{Xiaohong Liu}, \textbf{Yiran Qin}, \textbf{Philip Torr}, \textbf{Chen Zhang}, \textbf{Zhenfei Yin} \\[4pt]
{\normalsize Contact: \texttt{hengzzzhou@gmail.com}} \\[2pt]
{\normalsize \href{https://hengzzzhou.github.io/STS/}{\faGlobe\ \textsc{Project Page}} \quad \href{https://github.com/hengzzzhou/STS}{\faGithub\ \textsc{Code}}}
}

\begin{document}

\ifcolmsubmission
\linenumbers
\fi

\maketitle

\begin{abstract}
\input{sections/0_abstract.tex}
\end{abstract}

\input{sections/1_introduction.tex}
\input{sections/2_related_work.tex}
\input{sections/3_method.tex}

\input{sections/4_experiments.tex}
\input{sections/5_routing.tex}
\input{sections/5_discussion.tex}
\input{sections/6_conclusion.tex}

\clearpage

\bibliographystyle{colm2026_conference}
\bibliography{colm2026_conference}

\input{sections/appendix.tex}

\end{document}

%% file: sections/0_abstract.tex
When an LLM-based agent improves on a task, is the gain from the model or from the reasoning paradigm wrapped around it? We study this question by comparing six inference-time paradigms, namely Direct, CoT, ReAct, Plan-Execute, Reflection, and ReCode, across four frontier LLMs and ten benchmarks, yielding roughly 18k runs. We find that reasoning structure helps dramatically on some tasks but hurts on others: ReAct improves over Direct by 44pp on GAIA, while CoT degrades performance by 15pp on HumanEval. No single paradigm dominates, and oracle per-task selection beats the best fixed paradigm by 17.1pp on average. Motivated by this complementarity, we propose a \emph{select-then-solve} approach: before answering each task, a lightweight embedding-based router selects the most suitable paradigm. Across four models, the router improves average accuracy from 47.6\% to 53.1\%, outperforming the best fixed paradigm at 50.3\% by 2.8pp and recovering up to 37\% of the oracle gap. In contrast, zero-shot self-routing only works for GPT-5 at 67.1\% and fails for weaker models, all trailing the learned router. Our results argue that reasoning paradigm selection should be a per-task decision made by a learned router, not a fixed architectural choice.

%% file: sections/1_introduction.tex
\section{Introduction}\label{sec:introduction}

Reasoning paradigms have become a central design axis for LLM-based agents. A single large language model can be wrapped with different inference-time strategies: \emph{direct prompting} lets the model answer freely without imposing any reasoning scaffold, relying entirely on the model's own capabilities; \emph{chain-of-thought} explicitly instructs the model to reason step by step before answering~\citep{wei2022chain}; \emph{ReAct} interleaves reasoning with tool calls such as web search~\citep{yao2023react}; \emph{plan-then-execute} decomposes a task into a plan before acting~\citep{wang2023planandsolve}; \emph{reflection} generates an initial answer, critiques it, and revises~\citep{shinn2023reflexion,madaan2023selfrefine}; and \emph{ReCode} solves problems through recursive code generation and execution~\citep{wang2024recode,chen2023program}. These paradigms differ in how many LLM calls they make, whether they invoke tools, and how they allocate test-time computation~\citep{snell2024scaling}, yet they can all wrap the same base model.

Despite the growing variety of paradigms, controlled comparisons remain scarce. Prior work typically introduces a new paradigm and evaluates it on tasks tailored to its strengths, while changing the model, prompt format, tool stack, and benchmark simultaneously. As a result, the field has accumulated many positive case studies for individual paradigms, but much less evidence about a more basic question: \emph{when does additional reasoning structure actually help, and when does it hurt?}

We address this question by building \textsc{Paradigm}, a unified evaluation framework in which all six paradigms share the same model interface, evaluation code, and core tools, differing only in how they organize inference. We evaluate GPT-5~\citep{openai2025gpt5}, Gemini-3-Flash~\citep{google2025gemini3}, Qwen3-Max, and Qwen3-30B~\citep{qwen2025qwen3} on ten benchmarks covering code generation~\citep{chen2021humaneval}, mathematics~\citep{hendrycks2021math}, question answering~\citep{yang2018hotpotqa,kwiatkowski2019nq}, knowledge~\citep{hendrycks2021mmlu}, and tool-use-heavy tasks~\citep{mialon2023gaia,yao2024taubench}, yielding roughly 18k completed runs.

Our controlled comparison reveals that reasoning structure helps dramatically on some tasks but hurts on others. ReAct improves over Direct by 44pp on GAIA where web search is essential, while CoT degrades performance by 15pp on HumanEval where step-by-step reasoning disrupts code generation. The best paradigm changes systematically by dataset and by model: an oracle that selects the best paradigm per task outperforms the best fixed paradigm by 17.1pp on average.

This complementarity raises a natural question: if different tasks benefit from different paradigms, can we select the right paradigm before answering each task? We propose a \emph{select-then-solve} approach: a lightweight router analyzes the incoming task and dispatches it to the most suitable paradigm, including Direct itself when no structure is needed. Across four models, the router improves average accuracy from 47.6\% (Direct) to 53.1\%, outperforming the best fixed paradigm at 50.3\% by 2.8pp and recovering up to 37\% of the oracle gap. In contrast, zero-shot self-routing, where the model selects its own paradigm, only works for GPT-5 at 67.1\% while weaker models drop below their Direct baselines, revealing that reliable paradigm selection remains a challenging meta-reasoning capability.

We make the following contributions:\\[4pt]
\noindent$\diamond$~A controlled large-scale comparison of six inference-time paradigms across four LLMs and ten benchmarks.\\[2pt]
$\diamond$~Evidence that reasoning structure is helpful, neutral, or harmful depending on the task, with gains up to 44pp on information-seeking tasks and losses of 15pp on code generation.\\[2pt]
$\diamond$~A select-then-solve framework with an embedding-based paradigm router that improves average accuracy by 5.5pp over Direct and 2.8pp over the best fixed paradigm.\\[2pt]
$\diamond$~Evidence that self-routing is unreliable: only GPT-5 benefits while weaker models fail, all trailing the learned router.

\begin{figure}[t]
    \centering
    \includegraphics[width=\textwidth]{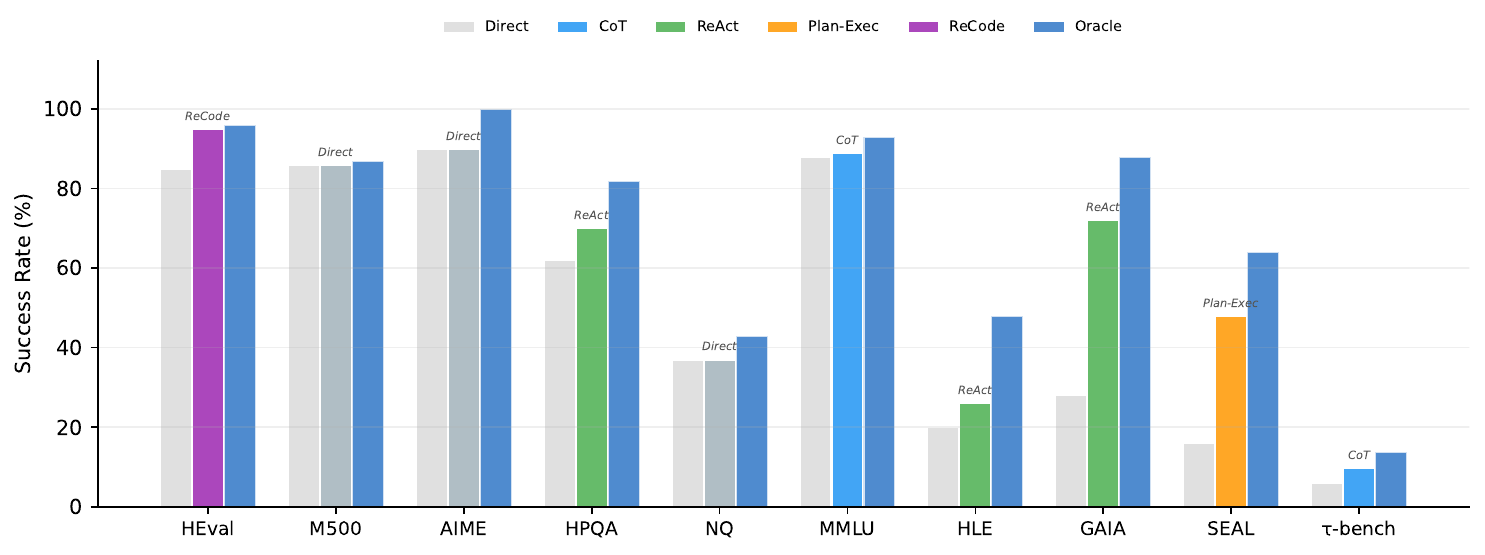}
    \caption{Direct prompting (gray), best single paradigm per dataset (colored), and oracle per-task selection (blue) for GPT-5. The best paradigm differs across tasks, and the oracle substantially exceeds any fixed choice, motivating our select-then-solve approach.}
    \label{fig:motivation}
\end{figure}

%% file: sections/2_related_work.tex
\section{Related Work}\label{sec:related_work}

\subsection{Inference-Time Reasoning Paradigms}

Chain-of-Thought prompting~\citep{wei2022chain} established that allocating generation budget to intermediate reasoning can improve performance on complex tasks, with later work exploring zero-shot CoT~\citep{kojima2022large}, self-consistency~\citep{wang2023selfconsistency}, and automatic rationale construction~\citep{zhang2023automatic}. Tool-augmented paradigms such as ReAct~\citep{yao2023react} extend this idea by interleaving reasoning with external actions, while planning-based methods separate decomposition from execution~\citep{wang2023planandsolve, liu2023llmp}. Self-refinement methods~\citep{madaan2023selfrefine, shinn2023reflexion} spend extra inference-time budget on critique and revision, and code-centric approaches such as PAL, Program-of-Thoughts, and ReCode~\citep{gao2023pal, chen2023program, wang2024recode} use executable programs as a reasoning substrate. Recent work has also explored reinforcement learning and search-based strategies for inference-time reasoning~\citep{zhang2025landscape, fan2025ssrl}, as well as comprehensive surveys of agent capabilities including memory, tool learning, and planning~\citep{yang2026toward}. Most of these works validate a single paradigm family against some baselines on tasks tailored to the method's strengths. Our focus is complementary: we ask how these paradigm families compare under a common implementation and evaluation pipeline, and whether their differences can be understood as task-dependent reasoning paradigms rather than isolated method wins.

\subsection{Agent Evaluation and Routing}

Benchmark suites such as AgentBench~\citep{liu2023agentbench}, WebArena~\citep{zhou2023webarena}, and MINT~\citep{wang2024mint} evaluate complete agent systems in realistic environments. These benchmarks are valuable, but they confound paradigm choice with many other system decisions, including the underlying model, prompting strategy, tool interfaces, and environment design. Our work isolates one design axis, the inference-time paradigm, within a shared framework.

Our routing analysis builds on a growing line of work on LLM routing and adaptive inference. RouteLLM~\citep{ong2024routellm} trains routers to dispatch queries between strong and weak models based on query difficulty. FrugalGPT~\citep{chen2023frugalgpt} cascades through models of increasing cost until a satisfactory answer is found. \citet{shnitzer2023large} and \citet{lu2024routing} route among multiple LLMs using benchmark-derived performance profiles. More recently, \citet{zhang2025avengers} unite smaller LLMs via routing to challenge proprietary models, \citet{zhang2025beyond} optimize routing for cost-performance trade-offs, and \citet{yue2025masrouter} learn to route LLMs within multi-agent systems. These approaches select among \emph{models}; we select among \emph{reasoning paradigms} for a fixed model, which is a complementary and largely unexplored axis of test-time optimization.

%% file: sections/3_method.tex
\section{Experimental Framework}\label{sec:framework}

\subsection{Paradigms as Inference-Time Policies}
\label{subsec:framework}

We treat a reasoning paradigm as a structured inference strategy $\mathcal{P}$ that maps a task $q$ to a sequence of language-model calls, optional tool invocations, intermediate state updates, and a final answer $\hat{y}$. Under this view, paradigm choice determines how much test-time computation is spent on planning, acting, revising, or executing code. For a task $\tau = (q, y^*)$, each paradigm induces
\begin{equation}
    \hat{y} = \mathcal{P}(\text{LLM}, q, \mathcal{T}),
\end{equation}
where $\mathcal{T}$ denotes the available tool repertoire. We then evaluate correctness with a dataset-specific scoring function $\mathrm{eval}_d(\hat{y}, y^*)$.

This framing lets us compare paradigms as algorithms for allocating inference-time computation. Importantly, we do not claim that every paradigm differs only in verbal reasoning style. Some paradigms invoke tools and others do not. We treat this as part of the deployed paradigm family being studied: ReAct without tools is not the same reasoning paradigm as the ReAct systems commonly used in practice.

\subsection{Studied Paradigms}
\label{subsec:paradigms}

We evaluate six representative paradigms that differ along two key dimensions: the degree of external control imposed on the model's reasoning process, and whether the paradigm grants access to external tools. Table~\ref{tab:paradigms} summarizes the paradigms studied.

\begin{table}[t]
\centering
\small
\caption{Reasoning paradigms studied in the main paper, organized by the degree of reasoning control and tool access.}
\label{tab:paradigms}
\begin{tabular*}{\textwidth}{@{\extracolsep{\fill}}llcccl@{}}
\toprule
\textbf{Paradigm} & \textbf{Key Idea} & \textbf{Control} & \textbf{Tools} & \textbf{LM Calls} & \textbf{Budget} \\
\midrule
Direct & Free-form answer & None & No & 1 & --- \\
CoT & Step-by-step reasoning & Instructed & No & 1 & --- \\
ReAct & Thought--action loop & Orchestrated & Yes & 1--15 & $T_{\max}=15$ \\
Plan-Execute & Plan, then act & Orchestrated & Yes & 2--16 & no replanning \\
Reflection & Critique and revise & Orchestrated & Yes & 3--45 & $M_{\max}=3$ \\
ReCode & Code as reasoning & Substrate & Yes & 2--10 & hierarchical \\
\bottomrule
\end{tabular*}
\end{table}

\paragraph{Reasoning control.} The paradigms impose varying levels of external structure on the model's reasoning. \textbf{Direct} imposes no control: the model receives the task and answers freely, using whatever internal reasoning it deems appropriate. This is not equivalent to ``no reasoning''; a capable model may internally perform multi-step derivation, but the scaffolding does not prescribe how. \textbf{CoT} adds instruction-level control by explicitly prompting ``think step by step,'' which can help weaker models organize their reasoning but may constrain stronger models that already reason effectively. \textbf{ReAct}, \textbf{Plan-Execute}, and \textbf{Reflection} impose orchestration-level control by structuring inference into multi-turn loops (thought-action, plan-then-execute, or generate-critique-revise). \textbf{ReCode} replaces the reasoning substrate entirely, using code generation and execution rather than natural language as the medium for problem-solving.

\paragraph{Tool access.} Orthogonally, paradigms differ in whether they grant the model access to external tools. Direct and CoT operate without tools, relying entirely on parametric knowledge. The remaining four paradigms provide a web-search interface and a Python code-execution tool. This distinction is critical: tool access enables the model to retrieve information beyond its training data, which explains why tool-using paradigms dominate on information-seeking tasks (e.g., GAIA, SEAL) while Direct often suffices for knowledge-centric tasks (e.g., MMLU, NQ).

The repository also contains exploratory strategies outside this paper's scope; all quantitative results in the main text use only the six paradigms above.

\subsection{Framework, Benchmarks, and Models}
\label{subsec:setup}

All paradigms share a common \texttt{BaseAgent} implementation for model access, tracing, metric collection, and result serialization. Tool-using paradigms access the same two tools: a web-search interface and a Python code-execution tool. Our unit of comparison is the paradigm family, not a fully orthogonalized factorial design over reasoning and tool access; this matches how paradigms are typically deployed.

We evaluate on ten benchmarks: HumanEval~\citep{chen2021humaneval}, MATH500~\citep{hendrycks2021math}, AIME, HotpotQA~\citep{yang2018hotpotqa}, Natural Questions~\citep{kwiatkowski2019nq}, MMLU~\citep{hendrycks2021mmlu}, HLE~\citep{phan2025hle}, GAIA~\citep{mialon2023gaia}, $\tau$-bench~\citep{yao2024taubench}, and SEAL, covering code generation, mathematics, QA, knowledge, and tool-use tasks. Table~\ref{tab:datasets} in the Appendix lists details. For large benchmarks we evaluate a fixed random subset (seed 42); for smaller ones we use the full set, yielding 761 tasks per model-paradigm pair and roughly 18k completed runs.

We evaluate four frontier LLMs: GPT-5, Qwen3-30B-A3B, Qwen3-Max, and Gemini-3-Flash. For every task we record correctness, total tokens, LLM call count, tool call count, and wall-clock time. Correctness is dataset-specific: code tasks use execution-based evaluation, math tasks use numeric matching, multiple-choice tasks use option extraction, and QA tasks use normalized text overlap and LLM as judge.

%% file: sections/4_experiments.tex
\section{Experiments}\label{sec:experiments}

Our experiments address four questions in sequence. (1)~How much does reasoning structure add beyond Direct prompting on each task type? (2)~What is the computational cost of that structure? (3)~Does the value of structure change across models of different capability? (4)~How complementary are the paradigms at the individual task level? The answers to these questions motivate the routing approach in Section~\ref{sec:routing}.

\subsection{Experimental Setup}
\label{subsec:setup}

The experimental code uses a unified \texttt{BaseAgent} with swappable strategy modules and a common result format. Each task result is cached as a JSON file keyed by (model, paradigm, dataset, task\_id), which makes interrupted runs resumable and lets us derive all summary tables from the raw per-task outputs. The repository also contains exploratory strategies beyond the six paradigms studied here; all numbers in this paper are produced from a paper-specific configuration released with the code, which defines an evaluation budget of 761 tasks per model-paradigm pair (18,264 runs in total, with roughly 18k completed runs in the current matrix).

Evaluation is dataset-specific. For HumanEval we extract Python code from model responses and execute against the benchmark tests. For MATH500 and AIME we extract numeric answers, including boxed and fractional \LaTeX{} forms, and compare with tolerance $\epsilon = 10^{-6}$. For HotpotQA, NQ, GAIA, and SEAL we apply normalized text matching with token-overlap fallback. For MMLU we extract the selected option letter. For $\tau$-bench we compare the predicted action sequence against the reference workflow. All prompts instruct the model to place its final answer inside \texttt{\textbackslash boxed\{\}}, ensuring consistent answer extraction across paradigms. Prompt templates are provided in the Appendix.

\subsection{The Marginal Value of Reasoning Structure}
\label{subsec:results}

Table~\ref{tab:full_results} presents the complete results across all four models, six paradigms, and ten datasets. We first analyze GPT-5, our strongest model, before examining cross-model patterns. The key question is whether reasoning structure provides consistent gains or whether its value depends on the task.

The marginal value of reasoning structure varies dramatically across tasks. Where tasks create information gaps that tools can fill, structure provides transformative gains: ReAct outperforms Direct by 44pp on GAIA and Plan-Execute improves 32pp on SEAL. In contrast, on knowledge-centric tasks where parametric knowledge suffices, Direct nearly matches the best paradigm: 88\% vs 89\% on MMLU, and best or tied-best on NQ and MATH500. Adding structure to these tasks increases cost without proportionate accuracy gain.

Structure can also actively hurt. CoT underperforms Direct by 15pp on HumanEval because step-by-step reasoning degrades code quality. At the other extreme, some tasks remain hard for all paradigms: the best paradigm achieves only 26\% on HLE and below 10\% on $\tau$-bench, demonstrating capability ceilings that no reasoning structure can overcome.

A striking pattern emerges: the \best{best} paradigm is different for nearly every dataset. CoT/ReAct leads on MMLU, ReAct on GAIA and HotpotQA, Plan-Execute on SEAL, ReCode on MATH500, and Direct on NQ and AIME. This raises a natural question: if different tasks favor different paradigms, what would happen if we could select the right paradigm for each task?

\subsection{The Cost of Reasoning Structure}
\label{subsec:cost}

Accuracy gains must be weighed against computational cost (Figure~\ref{fig:cost}, Appendix). Token usage forms clear tiers: Direct and CoT at 1.0--1.1$\times$, ReCode at 2.8$\times$, ReAct at 4.0$\times$, Plan-Execute at 6.9$\times$, and Reflection at 9.4$\times$. Reflection costs 9.4$\times$ more than Direct on HLE for only 2pp gain, while ReAct's overhead buys 44pp on GAIA.

\subsection{Structure as Capability Compensation}
\label{subsec:crossmodel}

If reasoning structure provides universally additive value, its contribution $\Delta_{\mathcal{P}}$ should be roughly constant across models. If it compensates for capability gaps, $\Delta_{\mathcal{P}}$ should be larger for weaker models. 

\begin{table}[t]
    \centering
    \caption{Complete success rates (\%) across all models, paradigms, and datasets. \best{Best} and \second{second best} paradigm per (model, dataset) are highlighted. Avg is the unweighted mean across all ten datasets.}
    \label{tab:full_results}
    \scriptsize
    \begin{tabular*}{\textwidth}{@{\extracolsep{\fill}}clccccccccccr@{}}
        \toprule
        \textbf{Model} & \textbf{Paradigm} & \textbf{HEval} & \textbf{M500} & \textbf{AIME} & \textbf{HPQA} & \textbf{NQ} & \textbf{MMLU} & \textbf{HLE} & \textbf{GAIA} & \textbf{SEAL} & \textbf{$\tau$-b} & \textbf{Avg} \\
        \midrule
        \multirow{6}{*}{\rotatebox{90}{GPT-5}} & Direct & 85.0 & \best{86.0} & \second{90.0} & 62.0 & \best{37.0} & \second{88.0} & 20.0 & 28.0 & 16.0 & 5.9 & 51.8 \\
         & CoT & 70.0 & \second{84.0} & 86.7 & 58.0 & 26.0 & \best{89.0} & 16.0 & 40.0 & 34.0 & \best{9.8} & 51.3 \\
         & ReAct & \second{86.0} & 80.0 & 86.7 & \best{70.0} & 27.0 & \best{89.0} & \best{26.0} & \best{72.0} & 28.0 & 5.9 & 57.1 \\
         & Plan-Exec & 35.0 & 75.0 & \second{90.0} & \second{69.0} & 27.0 & 83.0 & \second{22.0} & \second{68.0} & \best{48.0} & \second{7.8} & 52.5 \\
         & Reflect. & 85.0 & 69.0 & 75.0 & 67.0 & 29.0 & 85.0 & \second{22.0} & 58.0 & \second{46.0} & 5.9 & 54.2 \\
         & ReCode & \best{95.0} & \best{86.0} & 88.3 & 61.0 & \second{35.0} & 84.0 & 20.0 & 38.0 & 10.0 & \second{7.8} & 52.5 \\
        \midrule
        \multirow{6}{*}{\rotatebox{90}{Gemini-Flash}} & Direct & 73.0 & \best{82.0} & \second{40.0} & 64.0 & \best{43.0} & \best{90.0} & 14.0 & 44.0 & 22.0 & 7.8 & 48.0 \\
         & CoT & \best{99.0} & 76.0 & \best{53.3} & 62.0 & 30.0 & 60.0 & \second{16.0} & 54.0 & \best{62.0} & 3.9 & 51.6 \\
         & ReAct & \second{98.0} & 12.0 & 3.3 & \second{68.0} & 31.0 & \second{88.0} & 10.0 & \second{70.0} & \best{62.0} & 5.9 & 44.8 \\
         & Plan-Exec & \second{98.0} & 25.0 & 10.0 & \best{72.0} & 29.0 & 56.0 & 12.8 & \best{71.4} & 35.0 & \best{15.0} & 42.4 \\
         & Reflect. & \best{99.0} & 15.0 & 3.3 & 67.0 & 31.0 & 86.0 & 10.0 & 60.0 & \second{50.0} & \second{10.0} & 43.1 \\
         & ReCode & 81.0 & \second{80.0} & 26.7 & 64.0 & \second{39.0} & \best{90.0} & \best{25.0} & 45.0 & 25.0 & \best{15.0} & 49.1 \\
        \midrule
        \multirow{6}{*}{\rotatebox{90}{Qwen3-Max}} & Direct & 85.0 & \second{56.0} & 58.3 & 52.0 & 27.0 & \best{83.0} & \second{14.0} & 18.0 & 10.0 & \best{9.8} & 41.3 \\
         & CoT & \second{94.0} & 41.0 & 38.3 & \second{59.0} & 26.0 & 53.0 & \best{16.0} & 44.0 & \best{40.0} & 5.9 & 41.7 \\
         & ReAct & \best{96.0} & 50.0 & \second{61.7} & \second{59.0} & 28.0 & 66.0 & \best{16.0} & \best{56.0} & 30.0 & 2.0 & 46.5 \\
         & Plan-Exec & 71.0 & 47.0 & 55.0 & \best{65.0} & \second{29.0} & 68.0 & \second{14.0} & \second{48.0} & \second{38.0} & 0.0 & 43.5 \\
         & Reflect. & 92.0 & 46.0 & \second{61.7} & 54.0 & 27.0 & 10.0 & \second{14.0} & \second{48.0} & 30.0 & 2.0 & 38.5 \\
         & ReCode & \best{96.0} & \best{45.0} & 1.7 & 50.0 & \best{33.0} & 80.0 & 12.0 & 8.0 & 2.0 & 5.9 & 33.4 \\
        \midrule
        \multirow{6}{*}{\rotatebox{90}{Qwen3-30B}} & Direct & 18.0 & \second{42.0} & 10.0 & 27.0 & 20.0 & \second{70.0} & \best{14.0} & 12.0 & 6.0 & 0.0 & 21.9 \\
         & CoT & \second{64.0} & 15.0 & 3.3 & 34.0 & 19.0 & 48.0 & 2.0 & \best{28.0} & \best{28.0} & \best{3.9} & 24.5 \\
         & ReAct & 52.0 & 31.0 & 8.3 & \second{45.0} & 18.0 & 65.0 & 10.0 & \best{28.0} & 12.0 & 0.0 & 26.9 \\
         & Plan-Exec & 7.0 & 34.0 & \second{11.7} & 43.0 & \best{29.0} & 61.0 & \second{12.0} & \second{24.0} & \second{20.0} & 0.0 & 24.2 \\
         & Reflect. & 50.0 & 29.0 & 1.7 & \best{48.0} & \second{25.0} & 65.0 & 10.0 & 14.0 & \second{20.0} & 0.0 & 26.3 \\
         & ReCode & \best{96.0} & \best{33.0} & 5.0 & 17.0 & 17.0 & \best{70.0} & 8.0 & 4.0 & 0.0 & \best{3.9} & 25.4 \\
        \bottomrule
    \end{tabular*}
\end{table}

\paragraph{Reasoning structure compensates for capability gaps.} The data strongly support the compensation hypothesis. On HumanEval, CoT improves Qwen3-30B from 18\% to 64\%, a 46pp gain, while the same paradigm \emph{hurts} GPT-5 by 15pp. On GAIA, ReAct lifts Qwen3-30B from 12\% to 28\% and Gemini from 44\% to 70\%, but GPT-5 sees a similarly large jump from 28\% to 72\%. The marginal value of reasoning structure \emph{decreases} as model capability \emph{increases} on tasks within the model's reach, but remains large on tasks requiring external information regardless of model strength. On MATH500, the gap between Direct and the best paradigm narrows from 23pp for Qwen3-30B to 0pp for GPT-5, confirming that structure primarily compensates for what the model cannot do alone.

\paragraph{Task-type interactions are model-invariant.} Despite substantial variation in absolute performance, GPT-5 reaches 90.0\% on AIME while Qwen3-30B reaches only 11.7\%, the relative paradigm rankings remain remarkably stable across models. ReAct or Plan-Execute consistently tops GAIA and HotpotQA across all four models, ReCode leads on HumanEval for all models after our format fix, and Direct or ReCode dominates MATH500 universally. HLE remains hard at 14--26\% and $\tau$-bench near-zero for all. This stability suggests that task structure, not model capability, is the primary driver of which paradigm works best.

\subsection{Paradigm Complementarity}
\label{subsec:complementarity}

The preceding analysis repeatedly highlights one observation: the best paradigm differs across tasks. GAIA favors ReAct, MATH500 favors ReCode, SEAL favors Plan-Execute, and NQ favors Direct. But this is at the dataset level. We now ask: does this complementarity hold at the individual task level? If so, how large is the potential gain from task-level paradigm selection?

\paragraph{Oracle analysis.} For each task, we identify whether any of the five paradigms (excluding the Direct baseline) solves it correctly, and compute the oracle accuracy: the success rate achieved by always selecting the best paradigm per task.

\begin{table}[t]
    \centering
    \caption{Oracle analysis: per-task best paradigm selection achieves substantially higher accuracy than any single paradigm. Oracle Gap = Oracle $-$ Best-single.}
    \label{tab:oracle}
    \small
    \begin{tabular*}{\textwidth}{@{\extracolsep{\fill}}lcccc@{}}
        \toprule
        \textbf{Model} & \textbf{Direct} & \textbf{Best-Single} & \textbf{Oracle} & \textbf{Oracle Gap} \\
        \midrule
        GPT-5 & 60.3\% & 61.8\% (ReAct) & 74.1\% & +12.4pp \\
        Gemini-Flash & 48.0\% & 56.1\% (CoT) & 75.2\% & +19.1pp \\
        Qwen3-Max & 41.3\% & 46.5\% (ReAct) & 72.5\% & +21.6pp \\
        Qwen3-30B & 21.9\% & 31.7\% (ReAct) & 55.7\% & +24.1pp \\
        \midrule
        \textbf{Average} & 42.9\% & 49.0\% & 69.4\% & +17.1pp \\
        \bottomrule
    \end{tabular*}
\end{table}

Table~\ref{tab:oracle} reveals that oracle selection outperforms the best single paradigm by 17.1pp on average. The gap is largest for Qwen3-30B at 24.1pp and smallest for GPT-5 at 12.4pp, yet even for GPT-5 it represents a 20\% relative improvement.

\paragraph{Paradigms solve genuinely different problems.} This oracle gap could arise trivially if one paradigm solved almost all solvable tasks. Figure~\ref{fig:overlap} tests this by computing the Jaccard similarity between paradigm success sets. The relatively low overlap between ReCode and other paradigms confirms genuine complementarity. CoT and Reflection show the highest overlap at 0.61, yet even they disagree on 39\% of their combined success sets.

\paragraph{Summary.} Across four models and ten benchmarks, we find that reasoning structure is task-conditional, not universally beneficial. Structure helps most when it closes information gaps via tools, hurts when it constrains an already-capable model, and has no effect at capability ceilings. The 17.1pp oracle gap and low inter-paradigm overlap confirm that this variation is not noise but genuine complementarity. This motivates the routing approach we develop next.

%% file: sections/5_routing.tex
\section{Select-then-Solve: Paradigm Routing}\label{sec:routing}

\begin{figure}[b]
    \centering
    \includegraphics[width=\textwidth]{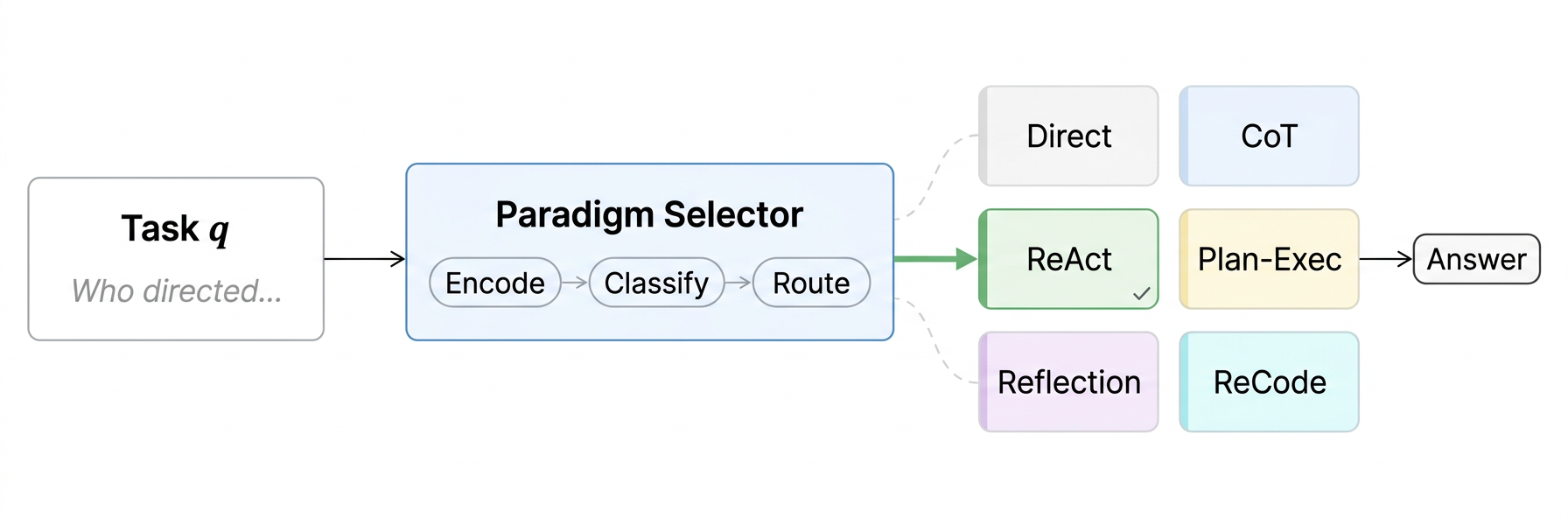}
    \caption{The select-then-solve pipeline. A Paradigm Selector encodes, classifies, and routes each task to one of six paradigms. Only the selected paradigm is executed.}
    \label{fig:pipeline}
\end{figure}

The 17.1pp oracle gap establishes that paradigm complementarity is substantial. A natural idea follows: before answering a task, first decide which reasoning paradigm to use. If the selection is good, the agent solves more tasks than any fixed paradigm choice.

\subsection{Routing Strategies}

We evaluate three routing strategies of increasing sophistication. The simplest uses 22 handcrafted features capturing dataset identity, text statistics, and content detectors, trained with Logistic Regression and a 2-layer MLP. The second encodes task text via \texttt{text-embedding-3-small} into 1536 dimensions, optionally concatenated with the handcrafted features. The third requires no training: each model reads its own task and selects a paradigm via zero-shot prompting. All routers select from six paradigms including Direct, so they can learn when no structure is needed. We train per-model classifiers on 70\% of tasks and evaluate on the held-out 30\%, measuring downstream task accuracy. Labels are the most token-efficient successful paradigm per task; tasks where all paradigms fail are excluded from training but counted as incorrect during evaluation.

\subsection{Routing Results}

\begin{figure}[t]
    \centering
    \includegraphics[width=\textwidth]{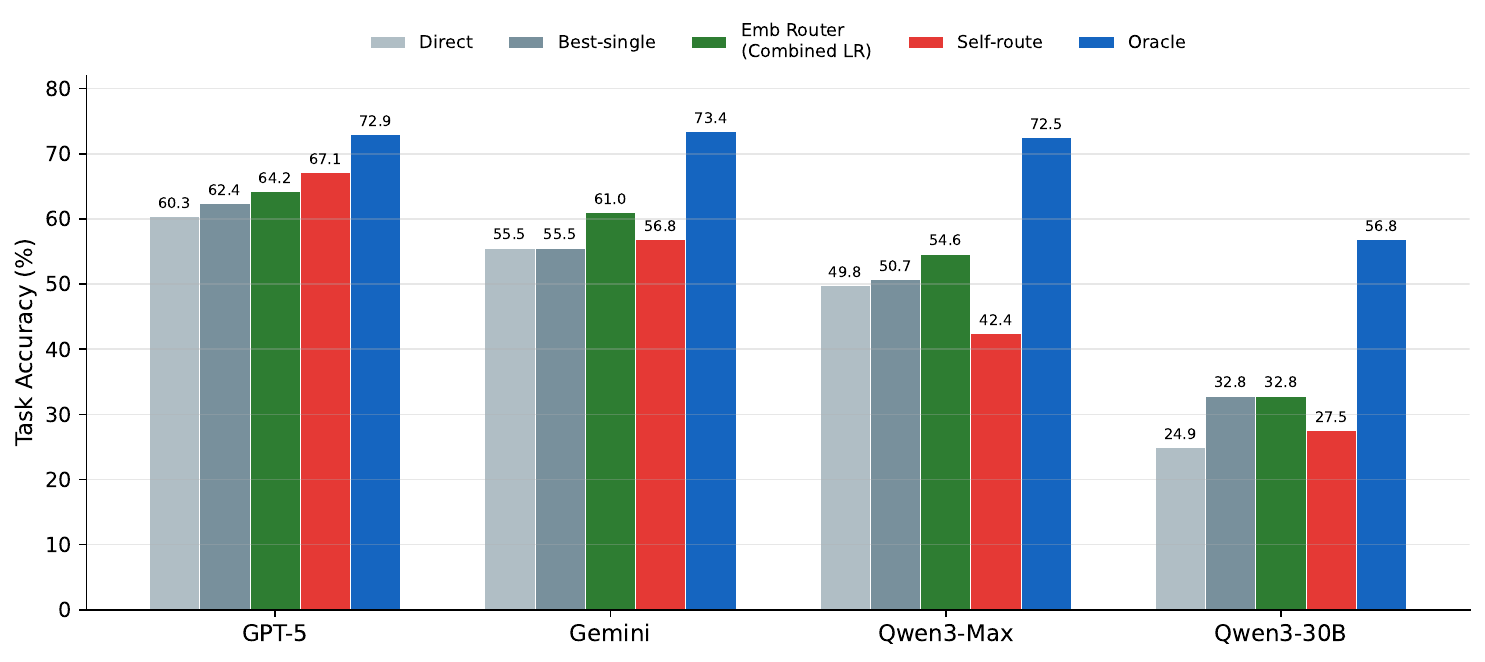}
    \caption{Router comparison across four models. The embedding router (green) consistently outperforms Direct and Best-single baselines. Self-routing (red) shows mixed results. The oracle (blue) shows substantial remaining headroom.}
    \label{fig:router_comparison}
\end{figure}

Figure~\ref{fig:router_comparison} and Table~\ref{tab:predictor} in the Appendix show a clear progression from simple to effective routing. The Combined LR router, which concatenates embeddings with handcrafted features, achieves 53.1\% average accuracy, improving 5.5pp over Direct at 47.6\% and 2.8pp over the best fixed paradigm at 50.3\%. Every embedding-based router exceeds Direct on all four models, with gains largest on Gemini at +6.7pp and GPT-5 at +3.9pp.

The upgrade from handcrafted to embedding features is meaningful. Handcrafted predictors rely on dataset identity, amounting to per-dataset majority voting. Embedding-based routers use the task text directly, enabling within-dataset discrimination. On GPT-5, the best handcrafted router merely ties the best fixed paradigm at 62.4\%, while the embedding router exceeds it at 64.2\%. On Gemini, the embedding LR at 61.9\% outperforms the handcrafted LR at 58.5\% by 3.4pp. Combined LR recovers 26\% of the oracle gap on average, reaching 31\% on GPT-5 and 37\% on Gemini, from a single embedding API call per task. Oracle gap recovery is computed as (Router $-$ Direct) / (Oracle $-$ Direct); for GPT-5: $(64.2 - 60.3)/(72.9 - 60.3) \approx 31\%$; for Gemini: $(62.2 - 55.5)/(73.4 - 55.5) \approx 37\%$. Because the router often selects Direct over expensive paradigms, the average token cost under routing is approximately 7.0k per task for GPT-5, compared to 3.5k for always-Direct and 14.0k for always-ReAct. The router achieves higher accuracy than ReAct at half its token cost.

\subsection{Self-Routing}

Zero-shot self-routing, where each model selects its own paradigm without training, averages 48.4\%, slightly above Direct at 47.6\%. GPT-5 achieves 67.1\%, surpassing its Direct baseline, but weaker models fail: Qwen3-Max drops to 42.4\% and Qwen3-30B to 27.5\%, well below their baselines. Weaker models overwhelmingly select tool-using paradigms regardless of task type, even for knowledge-centric questions where Direct suffices.

Table~\ref{tab:router_dist} reveals distribution asymmetry in detail. The learned router produces model-adapted distributions: it assigns 47--73\% of tasks to Direct depending on model capability, with the remainder spread across CoT, ReAct, and other paradigms. In contrast, self-routing collapses to a single dominant choice per model. GPT-5 over-selects Direct at 65\%, Qwen3-Max and Qwen3-30B over-select ReAct at 42--48\%, and no model ever selects Reflection. This pattern suggests that self-routing fails not because models lack knowledge of the paradigms, but because they cannot calibrate which paradigm matches each task's requirements. The ability to execute a paradigm when instructed does not imply the ability to choose when to use it, highlighting paradigm selection as a distinct meta-reasoning capability.

%% file: sections/5_discussion.tex
\section{Discussion}\label{sec:discussion}

\paragraph{What the router learns.}
Table~\ref{tab:router_dist} shows the paradigm distribution predicted by the Combined LR router. The router assigns Direct to 47--73\% of tasks depending on the model, confirming that it learns to avoid unnecessary reasoning structure. The fraction routed to Direct increases with model capability, from 47\% for Qwen3-30B to 66\% for GPT-5, consistent with our finding that stronger models need less structure. Weaker models receive more diverse routing across all six paradigms.

\begin{table}[t]
    \centering
    \caption{Paradigm distribution (\%) predicted by the learned router vs.\ zero-shot self-routing on the test set. Self-routing shows degenerate biases; the learned router produces diverse, model-adapted distributions.}
    \label{tab:router_dist}
    \scriptsize
    \begin{tabular*}{\textwidth}{@{\extracolsep{\fill}}llcccccc@{}}
        \toprule
        & \textbf{Model} & \textbf{Direct} & \textbf{CoT} & \textbf{ReAct} & \textbf{Plan-Exec} & \textbf{ReCode} & \textbf{Reflect.} \\
        \midrule
        \multirow{4}{*}{\rotatebox{90}{\scriptsize Router}} & GPT-5 & 65.9 & 9.6 & 14.4 & 7.9 & 1.3 & 0.9 \\
        & Gemini & 73.4 & 13.5 & 6.5 & 0.4 & 6.1 & 0.0 \\
        & Qwen3-Max & 58.9 & 19.7 & 2.6 & 7.0 & 10.9 & 0.9 \\
        & Qwen3-30B & 47.2 & 26.2 & 15.7 & 3.1 & 6.6 & 1.3 \\
        \midrule
        \multirow{4}{*}{\rotatebox{90}{\scriptsize Self}} & GPT-5 & 65.1 & 18.8 & 15.3 & 0.4 & 0.4 & 0.0 \\
        & Gemini & 72.1 & 10.9 & 16.6 & 0.0 & 0.4 & 0.0 \\
        & Qwen3-Max & 21.4 & 10.9 & 42.4 & 3.5 & 21.8 & 0.0 \\
        & Qwen3-30B & 16.6 & 29.3 & 48.0 & 0.9 & 5.2 & 0.0 \\
        \bottomrule
    \end{tabular*}
\end{table}

The contrast with self-routing is revealing. GPT-5 over-selects Direct at 65\%, achieving 67.1\% accuracy that exceeds Direct but still trails the learned router. Qwen3-Max and Qwen3-30B exhibit the opposite bias, over-selecting ReAct at 42--48\%, which hurts on knowledge-centric tasks. No model ever selects Reflection. While GPT-5's self-routing partially works, weaker models cannot calibrate their own paradigm selection.

\paragraph{When does Direct outperform structured paradigms?}
Direct prompting sometimes matches or exceeds more elaborate paradigms. This is not a confound but a central finding. We identify three scenarios: (1)~\emph{Knowledge-centric tasks}: on MMLU the answer exists in parametric knowledge and additional steps add latency without improving accuracy. (2)~\emph{Strong models on moderate tasks}: on MATH500, GPT-5 achieves 86\% with Direct, matching ReCode. (3)~\emph{Tasks where structure introduces errors}: on HumanEval, CoT underperforms Direct by 15pp because step-by-step reasoning degrades code quality. The router correctly learns these patterns: it routes 66\% of GPT-5 tasks to Direct, reserving structured paradigms for tasks where they provide genuine value.

%% file: sections/6_conclusion.tex
\section{Conclusion}\label{sec:conclusion}

We presented a controlled study of six inference-time reasoning paradigms across four LLMs and ten benchmarks. The value of reasoning structure is sharply task-dependent, creating complementarity that an embedding-based router exploits to outperform any fixed paradigm, recovering up to 37\% of the oracle gap at half the token cost of always using ReAct. Self-routing only works for the strongest model, revealing paradigm selection as a distinct meta-reasoning capability. The most effective agents may not be those with the most elaborate scaffolds, but those that know when to use them and when to step aside. Limitations and future work are discussed in Appendix~\ref{app:limitations}.

%% file: sections/appendix.tex
\appendix

\section{Limitations and Future Work}
\label{app:limitations}

\paragraph{Limitations.} We compare one implementation per paradigm family, not the full space of prompts, search budgets, or tool interfaces. We evaluate a fixed benchmark sample rather than re-sampling across seeds. Tool access is treated as part of the paradigm family rather than fully orthogonalized. Our router uses a modest training set of approximately 400 tasks per model; larger sets would likely improve accuracy.

\paragraph{Future work.} Immediate extensions include fine-tuning small language models as paradigm routers, multi-objective routing trading accuracy against cost, and transferring routing across models. More broadly, our framework supports adaptive strategies that mix paradigms within a single task, escalating from Direct to tool-using paradigms only when needed.

\section{Evaluation Benchmarks}

\begin{table}[h]
\centering
\small
\caption{Evaluation sets used in the paper. ``Eval instances'' refers to the number of tasks actually run per model-paradigm pair.}
\label{tab:datasets}
\begin{tabular*}{\textwidth}{@{\extracolsep{\fill}}llcll@{}}
\toprule
\textbf{Benchmark} & \textbf{Domain} & \textbf{Eval Instances} & \textbf{Metric} & \textbf{Notes} \\
\midrule
HumanEval & Code & 100 & Execution & function completion \\
MATH500 & Math & 100 & Numeric & multi-step derivation \\
AIME & Math & 60 & Numeric & olympiad-style problems \\
HotpotQA & QA & 100 & Text match & multi-hop QA \\
NQ & QA & 100 & Text match & factoid QA \\
MMLU & Knowledge & 100 & Letter match & broad subject coverage \\
HLE & Hard reasoning & 50 & LLM-as-judge & expert-level questions \\
GAIA & Agent QA & 50 & Text match & web/tool use required \\
$\tau$-bench & Tool planning & 51 & Action match & service workflows \\
SEAL & Current knowledge & 50 & Text match & recent-event QA \\
\bottomrule
\end{tabular*}
\end{table}

\section{Paradigm Prompt Templates}
\label{app:prompts}

All six paradigms share the same \texttt{BaseAgent} framework and differ only in the system prompts and control flow. We list the core prompts below. All prompts include the \texttt{\textbackslash boxed\{\}} instruction to ensure consistent answer extraction.

\paragraph{Direct.}
\begin{quote}
\small\ttfamily
You are a helpful assistant. Answer the question directly and concisely. Put your final answer inside \textbackslash boxed\{\}, e.g., \textbackslash boxed\{42\} or \textbackslash boxed\{Yes\}.
\end{quote}

\paragraph{Chain-of-Thought (CoT).}
\begin{quote}
\small\ttfamily
You are a helpful assistant. Think through the problem step by step before giving your final answer. Show your reasoning, then put your final answer inside \textbackslash boxed\{\}.
\end{quote}

\paragraph{ReAct.}
\begin{quote}
\small\ttfamily
You are a helpful assistant that can use tools to answer questions. For each step, you should: 1) Think about what you need to do next. 2) If needed, use a tool to gather information. 3) Once you have enough information, provide your final answer. Put your final answer inside \textbackslash boxed\{\}.
\end{quote}

\paragraph{Plan-then-Execute.} Uses two prompts: a planning prompt that produces a numbered step list, followed by an execution prompt:
\begin{quote}
\small\ttfamily
[Planning] First, create a step-by-step plan to solve the problem. Output ONLY the plan as a numbered list. Do not solve the problem yet.\\[4pt]
[Execution] You are executing a plan step by step. You have access to tools. After completing all steps, put your final answer inside \textbackslash boxed\{\}.
\end{quote}

\paragraph{Reflection.} Uses three phases: initial solve, critique, and revision.
\begin{quote}
\small\ttfamily
[Solve] Think carefully and use tools when needed to find the answer. Put your final answer inside \textbackslash boxed\{\}.\\[4pt]
[Reflect] Examine the following answer and identify any errors. If the answer is correct, respond with ``SATISFACTORY''. If there are issues, explain what's wrong.\\[4pt]
[Revise] Your previous answer had issues. Revise based on the feedback. Put your final answer inside \textbackslash boxed\{\}.
\end{quote}

\paragraph{ReCode.} Generates Python code with placeholder functions, then decomposes recursively:
\begin{quote}
\small\ttfamily
[Generate] Write a Python function solve() that solves the problem. Mark placeholder functions with \# PLACEHOLDER. Available primitives: web\_search(query), code\_exec(code).\\[4pt]
[Decompose] Implement the placeholder function \{func\_name\}. You may create new placeholders if needed.\\[4pt]
[Extract] Given the execution output, extract the final answer. Put it inside \textbackslash boxed\{\}.
\end{quote}

\section{Detailed Paradigm Comparison}
\label{app:paradigm_comparison}

Table~\ref{tab:paradigm_detail} provides a detailed comparison of how the six paradigms differ in their inference-time behavior.

\begin{table}[h]
    \centering
    \caption{Detailed comparison of the six paradigms. LM calls and tool calls are per-task averages on GPT-5.}
    \label{tab:paradigm_detail}
    \scriptsize
    \setlength{\tabcolsep}{3pt}
    \begin{tabular}{lccccl}
        \toprule
        \textbf{Paradigm} & \textbf{Avg LM Calls} & \textbf{Avg Tool Calls} & \textbf{Tools?} & \textbf{Multi-turn?} & \textbf{Key Mechanism} \\
        \midrule
        Direct & 1.0 & 0 & No & No & Single-shot generation \\
        CoT & 1.0 & 0 & No & No & Step-by-step verbal reasoning \\
        ReAct & 1--15 & 0--14 & Yes & Yes & Thought-action-observation loop \\
        Plan-Exec & 2--16 & 0--15 & Yes & Yes & Plan first, then execute with tools \\
        Reflection & 3--45 & 0--42 & Yes & Yes & Solve, critique, revise (up to 3 rounds) \\
        ReCode & 2--10 & 0--8 & Yes & Yes & Recursive code decomposition \\
        \bottomrule
    \end{tabular}
\end{table}

The paradigms form a spectrum of increasing inference-time structure. Direct and CoT are single-call, zero-tool paradigms that differ only in whether step-by-step reasoning is elicited. ReAct and Plan-Execute both use tools but differ in control flow: ReAct interleaves reasoning and action in a flat loop, while Plan-Execute separates planning from execution. Reflection adds a self-critique phase after initial solving. ReCode is unique in using code generation as the primary reasoning substrate.

\section{Router Training Details}
\label{app:router_training}

\paragraph{Label construction.} The router's training labels are derived from the experimental matrix without any additional LLM calls. For each (model, task) pair, we observe the outcomes of all six paradigms from the main experiment. If at least one paradigm succeeds, the label is the successful paradigm with the lowest token cost (favoring cheaper paradigms when multiple succeed). If all paradigms fail, the task is labeled \texttt{none} and excluded from training but counted as incorrect during evaluation. This yields approximately 300--400 labeled training examples per model.

\paragraph{Features.} We compare three feature representations:
(1)~\emph{Handcrafted}: 22 features including dataset one-hot encoding (10 dims), text statistics (length, word count, line count, average word length), content flags (has\_code, has\_math, has\_choices), and question-type indicators (5 dims).
(2)~\emph{Embedding}: the task text is encoded via the \texttt{text-embedding-3-small} API, producing a 1536-dimensional vector.
(3)~\emph{Combined}: concatenation of the embedding and handcrafted features (1558 dims total).

\paragraph{Classifiers.} For each feature set, we train Logistic Regression (LR; \texttt{class\_weight=balanced}, \texttt{max\_iter=2000}) and a 2-layer MLP (128$\to$64$\to$6, dropout 0.3, Adam optimizer with learning rate $5 \times 10^{-4}$, early stopping with patience 15). Features are standardized to zero mean and unit variance before training. All classifiers are trained independently per model.

\paragraph{Train/test split.} Tasks are split 70/30 stratified by dataset, with the same split shared across all models (532 train / 229 test). The split is fixed with seed 42 for reproducibility.

\paragraph{Deployment cost.} At inference time, the router requires one embedding API call per task (negligible latency and cost) followed by a linear classifier prediction. No LLM calls are needed for routing. The total overhead is orders of magnitude cheaper than running a single paradigm, let alone all six.

\paragraph{LLM configuration.} All models are accessed through an OpenAI-compatible API with \texttt{temperature=0} for deterministic outputs. The maximum context length varies by model: GPT-5 supports 128k tokens, Gemini-3-Flash 1M tokens, and Qwen3 models 32k tokens. For tool-using paradigms, we set a maximum of 15 interaction turns for ReAct, 16 for Plan-Execute, and 3 revision rounds for Reflection. The code execution tool enforces a 30-second timeout per execution. All paradigms use the same system prompt format with \texttt{\textbackslash boxed\{\}} answer extraction. The embedding model \texttt{text-embedding-3-small} produces 1536-dimensional vectors and is called with default parameters. Complete prompt templates are listed in Appendix~\ref{app:prompts}.

\section{Detailed Router Results}

\begin{table}[h]
    \centering
    \caption{Downstream task accuracy (\%) on the held-out test set. Each router selects a paradigm per task (including Direct); accuracy is the fraction of tasks solved correctly under the selected paradigm. Best router per model is \textbf{bolded}.}
    \label{tab:predictor}
    \scriptsize
    \begin{tabular*}{\textwidth}{@{\extracolsep{\fill}}llccccc@{}}
        \toprule
        & \textbf{Method} & \textbf{GPT-5} & \textbf{Gemini} & \textbf{Qwen3-Max} & \textbf{Qwen3-30B} & \textbf{Avg} \\
        \midrule
        \multirow{2}{*}{\rotatebox{90}{\scriptsize Base}} & Direct & 60.3 & 55.5 & 49.8 & 24.9 & 47.6 \\
        & Best-single & 62.4 & 55.5 & 50.7 & 32.8 & 50.3 \\
        \cmidrule{1-7}
        \multirow{2}{*}{\rotatebox{90}{\scriptsize HC}} & Handcrafted (LR) & 59.4 & 58.5 & 51.5 & 34.9 & 51.1 \\
        & Handcrafted (MLP) & 62.4 & 55.5 & 53.3 & 37.1 & 52.1 \\
        \cmidrule{1-7}
        \multirow{2}{*}{\rotatebox{90}{\scriptsize Emb}} & Embedding (LR) & 64.2 & 61.9 & 53.3 & 32.3 & 52.9 \\
        & Combined (LR) & \textbf{64.2} & 61.0 & \textbf{54.6} & 32.8 & \textbf{53.1} \\
        & Combined (MLP) & 62.4 & \textbf{62.2} & 53.3 & \textbf{33.6} & 52.9 \\
        \cmidrule{1-7}
        \multirow{1}{*}{\rotatebox{90}{\scriptsize ZS}} & Self-route & 60.3 & 56.8 & 42.4 & 27.5 & 46.7 \\
        \cmidrule{1-7}
        & Oracle & 72.9 & 73.4 & 72.5 & 56.8 & 68.9 \\
        \bottomrule
    \end{tabular*}
\end{table}

\section{Additional Figures}

\begin{figure}[t]
    \centering
    \includegraphics[width=0.75\textwidth]{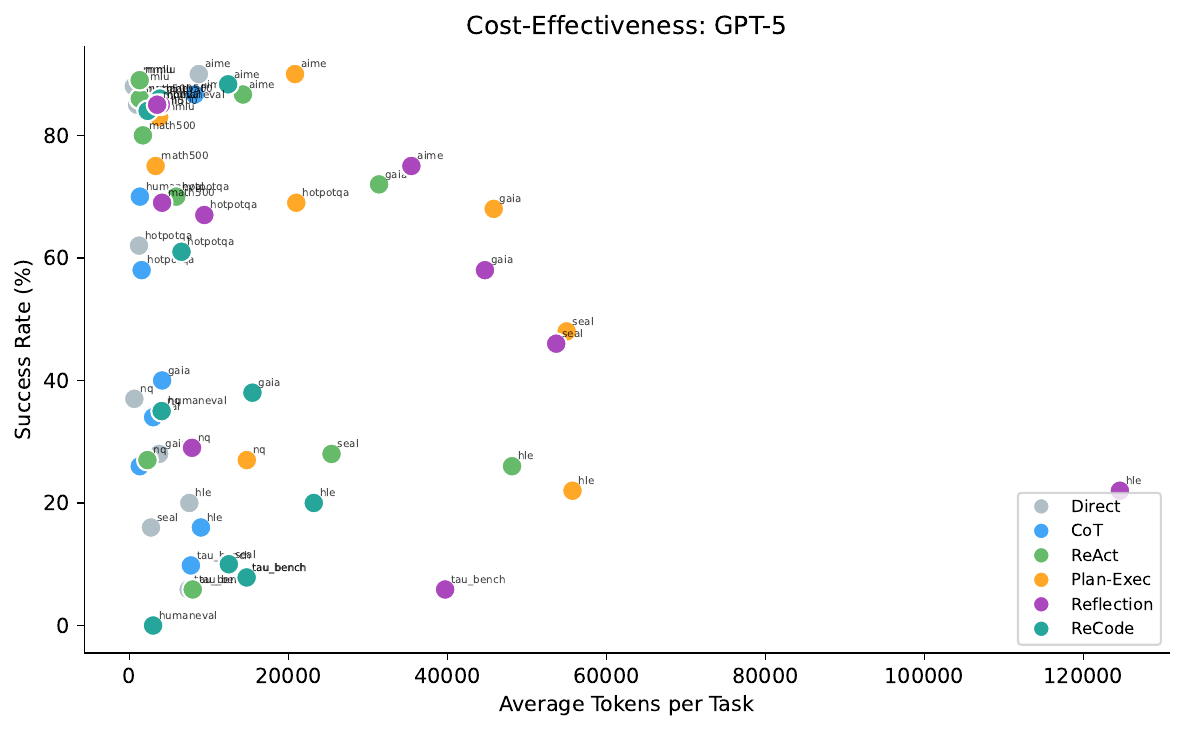}
    \caption{Cost-effectiveness scatter plot: success rate vs.\ average tokens per task. Points closer to the top-left corner represent better cost-efficiency.}
    \label{fig:cost}
\end{figure}

\begin{figure}[t]
    \centering
    \includegraphics[width=\textwidth]{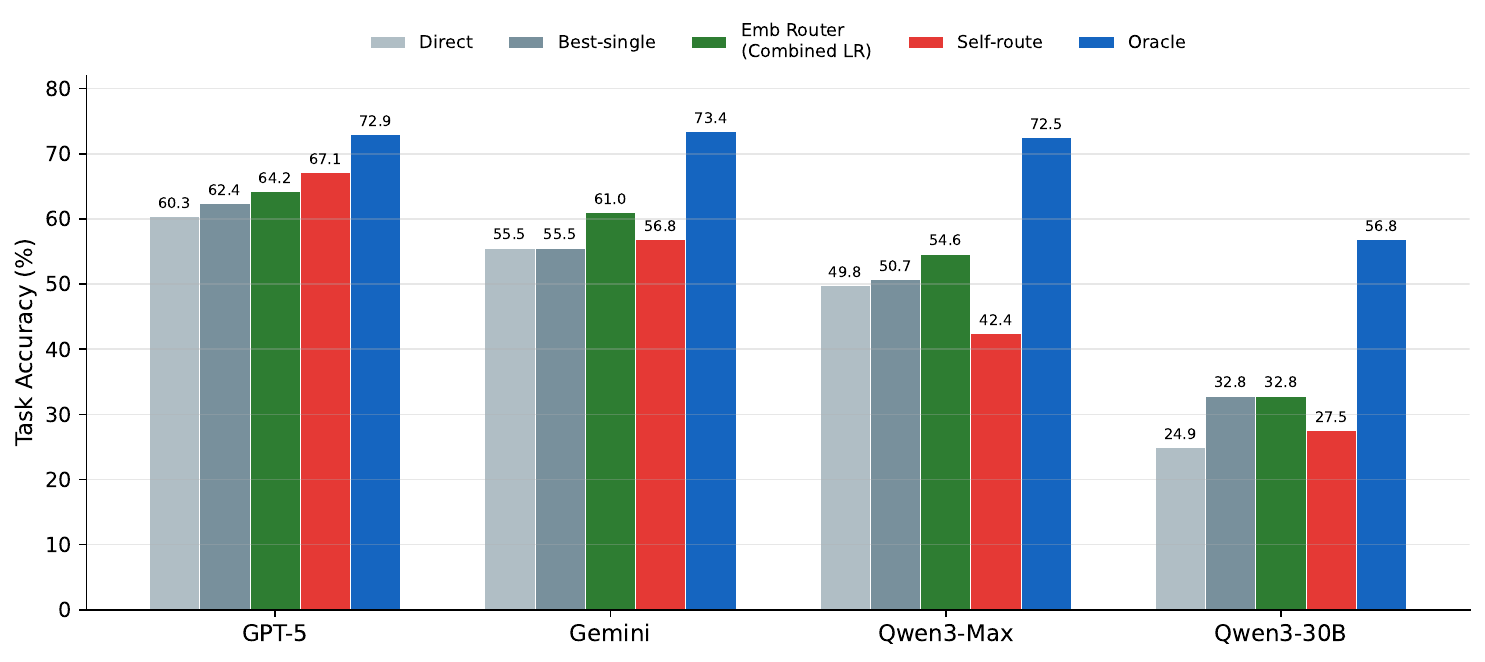}
    \caption{Router comparison across four models. The embedding router (green) consistently outperforms Direct and Best-single baselines. Self-routing (red) fails to improve over Direct on most models. The oracle (blue) shows substantial remaining headroom.}
    \label{fig:router_comparison_app}
\end{figure}

\begin{figure}[t]
    \centering
    \includegraphics[width=\textwidth]{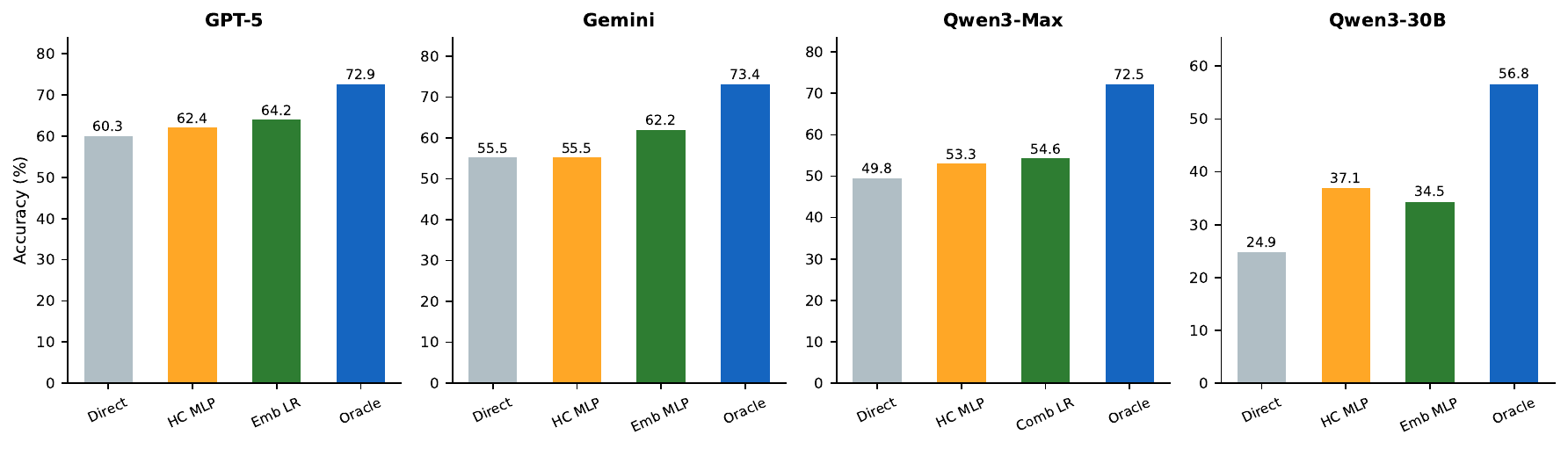}
    \caption{Oracle gap recovery across models. Each group shows the progression from Direct (gray) through handcrafted router (orange) and embedding router (green) to the Oracle upper bound (blue). The embedding router consistently narrows the gap, with the largest recovery on Gemini.}
    \label{fig:oracle_recovery}
\end{figure}

\begin{figure}[t]
    \centering
    \includegraphics[width=\textwidth]{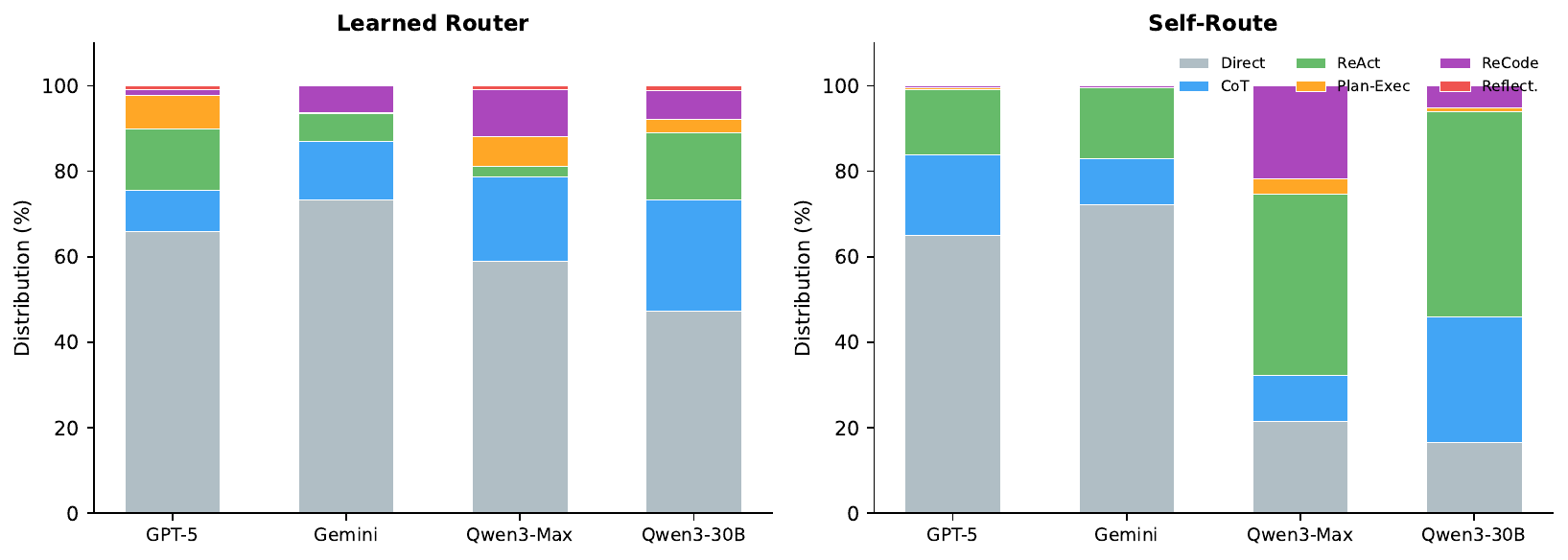}
    \caption{Paradigm distribution comparison: learned router (left) vs.\ zero-shot self-routing (right). The learned router produces diverse, model-adapted distributions, while self-routing shows degenerate biases (GPT-5 selects Direct 100\%; Qwen3-30B over-selects ReAct at 48\%).}
    \label{fig:paradigm_dist}
\end{figure}

\begin{figure}[t]
    \centering
    \includegraphics[width=0.9\textwidth]{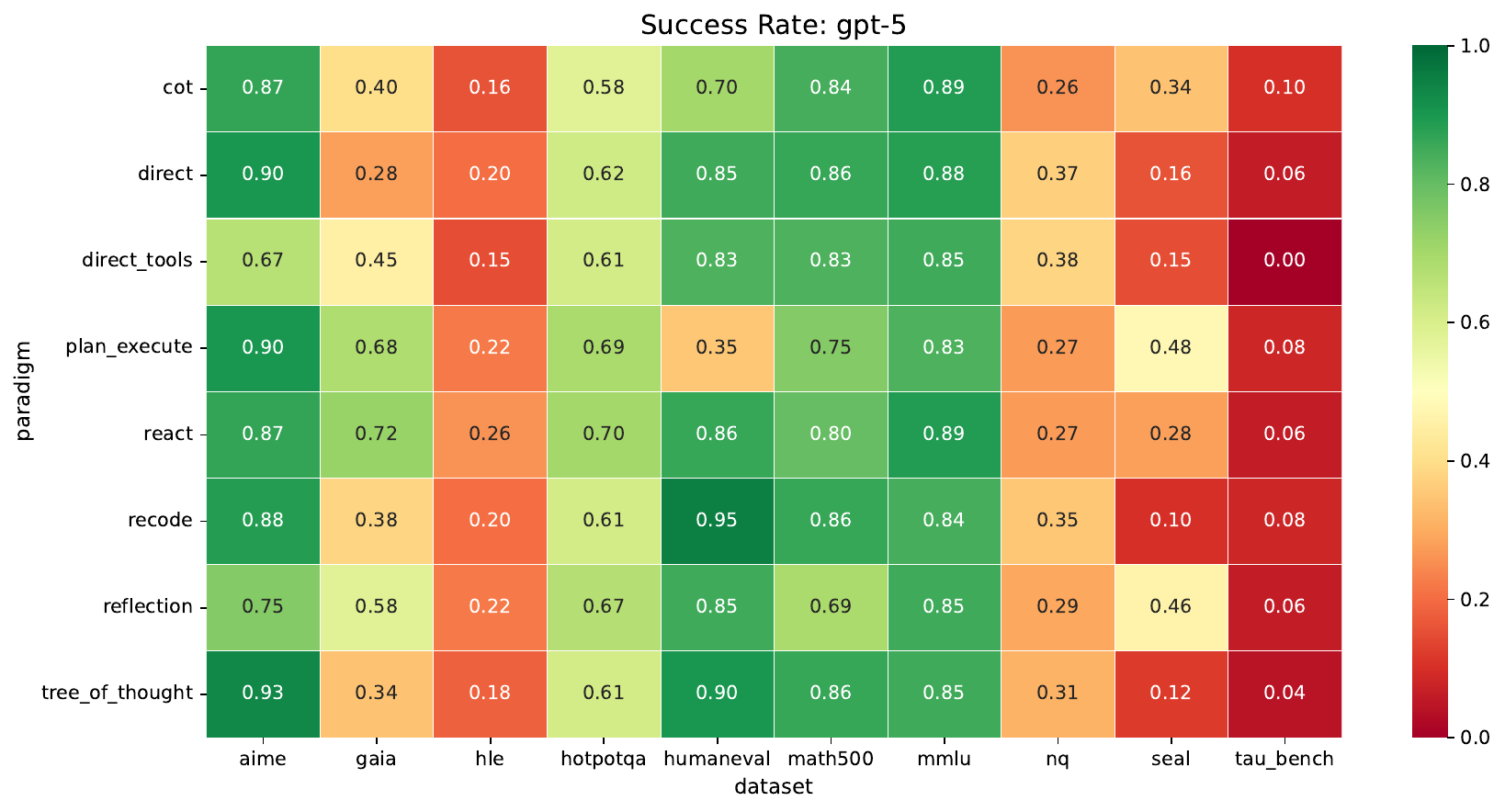}
    \caption{Success rate heatmap for GPT-5 across paradigms (rows) and datasets (columns). Clear task-paradigm interactions are visible: no single row dominates all columns.}
    \label{fig:heatmap_app}
\end{figure}

\begin{figure}[t]
    \centering
    \includegraphics[width=0.7\textwidth]{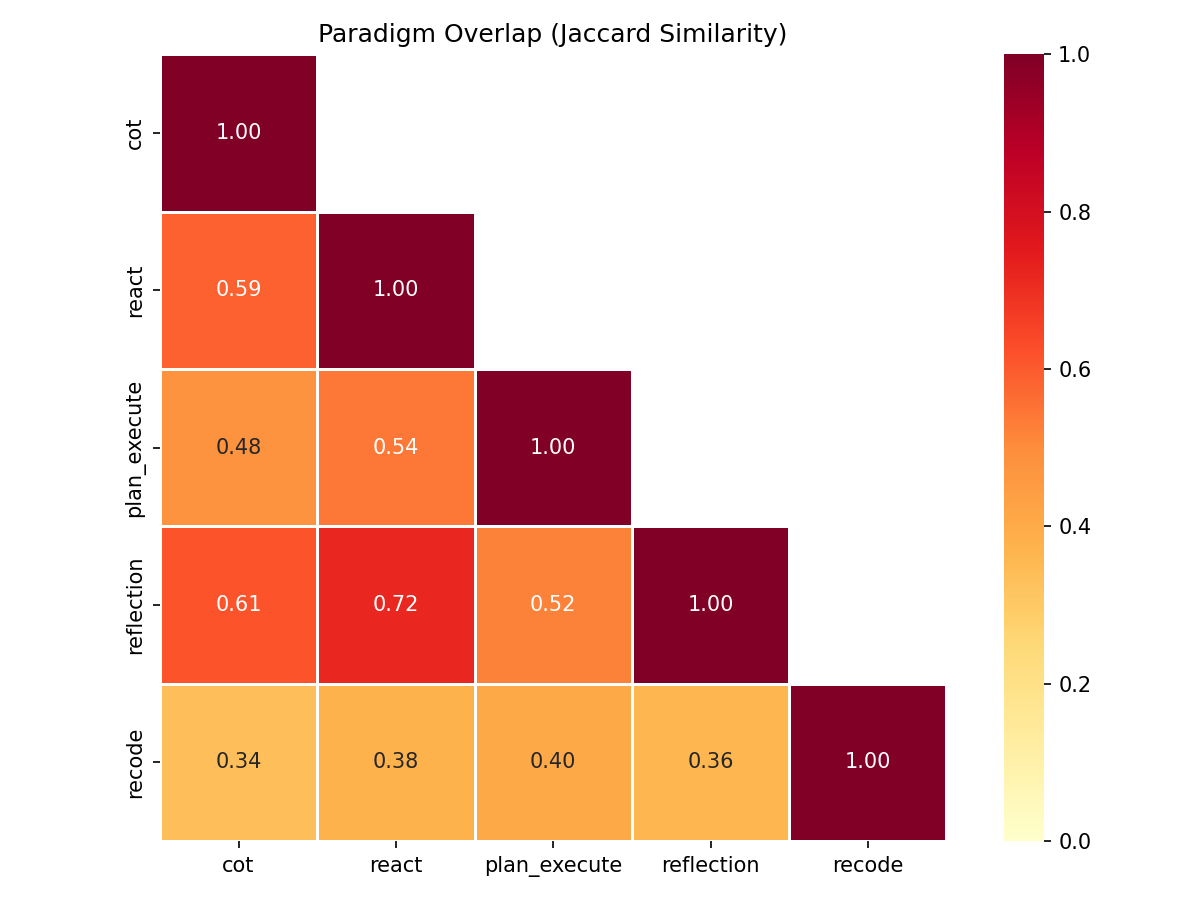}
    \caption{Jaccard similarity between paradigm success sets (aggregated across all models). Lower values indicate greater complementarity; paradigms solve different subsets of tasks.}
    \label{fig:overlap}
\end{figure}


\section{Case Studies}
\label{app:case_studies}

We present four representative cases from GPT-5 that illustrate the key patterns in our findings. For each case, we show the task, the outcome across paradigms, and the qualitative failure or success mode.

\paragraph{Case 1: Structure hurts (MMLU).}

\begin{quote}
\small \emph{A kidney dialysis center periodically checks equipment and recalibrates if readings are off target. A fabric factory checks towel sizes and halts production if measurements are off target. In both situations, the null hypothesis is that equipment performs satisfactorily. Which is more serious, a Type~I or Type~II error?} \\
\textbf{Answer: C} (Dialysis: Type~II; Towels: Type~I)
\end{quote}

\begin{center}
\scriptsize
\begin{tabular}{lccc}
\toprule
\textbf{Paradigm} & \textbf{Success} & \textbf{Answer} & \textbf{Tokens} \\
\midrule
Direct & \checkmark & C & 900 \\
CoT & \checkmark & C & 1,872 \\
ReAct & \ding{55} & D & 1,350 \\
Plan-Execute & \ding{55} & D & 1,878 \\
Reflection & \ding{55} & D & 3,064 \\
ReCode & \checkmark & C & 1,536 \\
\bottomrule
\end{tabular}
\end{center}

ReAct, Plan-Execute, and Reflection all answered D, arguing that Type~II error is more serious in \emph{both} cases. Their extended reasoning about the towel manufacturer led them to over-weight the consequences of missed defects, failing to recognize that for towels (a low-stakes product), an unnecessary production halt (Type~I) is the more costly concern. Reflection even validated its own wrong answer as ``SATISFACTORY'' during self-critique, illustrating how self-reflection can reinforce rather than correct an initial error. Direct, with minimal reasoning overhead, correctly identified C.

\paragraph{Case 2: Tools are essential (GAIA).}

\begin{quote}
\small \emph{Who did the actor who played Ray in the Polish-language version of Everybody Loves Raymond play in Magda M.? Give only the first name.} \\
\textbf{Answer: Wojciech}
\end{quote}

\begin{center}
\scriptsize
\begin{tabular}{lcccc}
\toprule
\textbf{Paradigm} & \textbf{Success} & \textbf{Answer} & \textbf{Tokens} & \textbf{Tool Calls} \\
\midrule
Direct & \ding{55} & Andrzej & 5,192 & 0 \\
CoT & \ding{55} & Bartek & 4,566 & 0 \\
ReAct & \checkmark & Wojciech & 21,248 & 7 \\
Plan-Execute & \checkmark & Wojciech & 15,417 & 5 \\
Reflection & \checkmark & Wojciech & 13,177 & 3 \\
ReCode & \ding{55} & Piotr & 1,889 & 1 \\
\bottomrule
\end{tabular}
\end{center}

This multi-hop question requires: (1) identifying the Polish adaptation ``Wszyscy Kochaja Romana'', (2) finding the actor who played the Ray counterpart (Bartlomiej Kasprzykowski), and (3) looking up his role in Magda~M. Without tools, every paradigm hallucinated a different Polish first name. The three paradigms with sufficient tool interaction (ReAct with 7 searches, Plan-Execute with 5, Reflection with 3) all converged on the correct answer. ReAct's trace shows it progressively refining its web queries from English to Polish-language searches before finding the correct actor and role.

\paragraph{Case 3: Paradigm complementarity (SEAL).}

\begin{quote}
\small \emph{As reported by the FAO, which nation ranked as the world's second-largest rice producer in 2023?} \\
\textbf{Answer: China}
\end{quote}

\begin{center}
\scriptsize
\begin{tabular}{lcccc}
\toprule
\textbf{Paradigm} & \textbf{Success} & \textbf{Answer} & \textbf{Tokens} & \textbf{Tool Calls} \\
\midrule
Direct & \ding{55} & India & 648 & 0 \\
CoT & \ding{55} & India & 966 & 0 \\
ReAct & \checkmark & China & 11,174 & 5 \\
Plan-Execute & \checkmark & China & 25,600 & 9 \\
Reflection & \ding{55} & India & 32,650 & 10 \\
ReCode & \ding{55} & India & 9,168 & 1 \\
\bottomrule
\end{tabular}
\end{center}

This question tests whether parametric knowledge can be overridden by evidence. The model's default belief strongly associates India as the \#2 rice producer, but the 2023 FAO data places China second. Five paradigms answered ``India'' based on parametric knowledge. Most strikingly, Reflection made 10 tool calls and \emph{still} answered ``India'': it found conflicting evidence but its self-reflection loop reinforced the initial (wrong) parametric belief. Only ReAct and Plan-Execute, which are structured to act on retrieved evidence rather than validate prior beliefs, answered correctly. This illustrates that tool access alone is insufficient; the paradigm must also be structured to \emph{trust} retrieved evidence over prior beliefs.

\paragraph{Case 4: Capability ceiling (HLE).}

\begin{quote}
\small \emph{The braid group $B_n$ acts on the torus link $T(n,n) \subset S^3$ by permuting strands, inducing an action on $Kh(T(n,n); \mathbb{Q})$. Let $d_n$ be the dimension of the $B_n$-fixed subspace. Find $\prod_{n=1}^8 d_n$.} \\
\textbf{Answer: 2,490,840,000}
\end{quote}

\begin{center}
\scriptsize
\begin{tabular}{lcccc}
\toprule
\textbf{Paradigm} & \textbf{Success} & \textbf{Answer} & \textbf{Tokens} & \textbf{Tool Calls} \\
\midrule
Direct & \ding{55} & 362,880 & 5,300 & 0 \\
CoT & \ding{55} & 362,880 & 7,849 & 0 \\
ReAct & \ding{55} & 362,880 & 86,430 & 15 \\
Plan-Execute & \ding{55} & 362,880 & 103,027 & 15 \\
Reflection & \ding{55} & 362,880 & 79,097 & 13 \\
ReCode & \ding{55} & 362,880 & 13,132 & 1 \\
\bottomrule
\end{tabular}
\end{center}

All six paradigms produce the same wrong answer: 362,880 ($= 9!$). Each conjectures $d_n = n+1$ via a plausible but incorrect argument citing Schur-Weyl duality and symmetric tensors. The actual computation requires detailed knowledge of Khovanov homology that exceeds the model's training data. ReAct spent 86,430 tokens and 15 tool calls searching for relevant references but could not locate the specific computations needed. The total cost across all paradigms for this single failed task was over 295,000 tokens. This demonstrates a hard capability ceiling where no amount of orchestration or tool use can compensate for missing domain knowledge.